\documentclass[runningheads]{llncs}
\usepackage{booktabs}
\usepackage[T1]{fontenc}
\usepackage{graphicx}
\usepackage{float}  
\begin{document}
\title{PKRD-CoT: A Unified Chain-of-thought Prompting for Multi-Modal Large Language Models in Autonomous Driving}
\author{Xuewen Luo\orcidID{0009-0006-7737-1178} \and
Fan Ding\orcidID{0009-0007-0233-4310} \and
Yinsheng Song\orcidID{0009-0001-1423-0275} \and
Xiaofeng Zhang\inst{*}\orcidID{0000-0002-7185-4682} \and
Junnyong Loo\inst{*}\orcidID{0000-0001-9370-600X}}

%

\institute{School of Information Technology, Monash University Malaysia, Bandar Sunway, 47500 Selangor, Malaysia
\email{xluo0033@student.monash.edu}}
\maketitle              % typeset the header of the contribution
\begin{abstract}
There are growing interest in leveraging the capabilities of robust Multi-Modal Large Language Models (MLLMs) directly within autonomous driving contexts. However, the high costs and complexity of designing and training end-to-end autonomous driving models make them difficult for many enterprises and research entities. To address this, our study explores a seamless integration of MLLMs into autonomous driving systems, by proposing a Zero-Shot Chain-of-Thought (Zero-shot-CoT) prompt design named PKRD-CoT. PKRD-CoT is constructed based on the four fundamental capabilities of autonomous driving—perception, knowledge, reasoning, and decision-making—making it particularly suitable for understanding and responding to dynamic driving environments by mimicking human thought processes step by step to enhance decision-making in real-time scenarios. Our design enables MLLMs to tackle problems without prior experience, thus enhancing its utility within unstructured autonomous driving environments. In experiments, we demonstrate the exceptional performance of GPT 4.0 with PKRD-CoT across autonomous driving tasks, highlighting its effectiveness for application in autonomous driving scenarios. Additionally, our benchmark analysis reveals promising viability of PKRD-CoT for other MLLMs such as Claude, LLava1.6, and Qwen-VL-Plus. Overall, this study contributes a novel and unified prompt designing framework for GPT 4.0 and other MLLMs in autonomous driving, while also evaluating the efficacy of these widely recognized MLLMs in the autonomous driving domain through rigorous comparisons.

\end{abstract}
\section{Introduction}

Previously, autonomous driving technologies were dominated by data-driven modelling \cite{bolte2019towards}, where deep learning advancements have allowed cars to "learn" how to drive from large amounts of driving data. However, such an approach require us to train the car on a large amount of data, giving rise to dilemmas such as dataset bias, overfitting, non-interpretability, and poor generalization ability \cite{bogdoll2021description}.

On the contrary, the nature of human driving skills is knowledge-driven rather than data-driven through the exploration of several studies \cite{wen2023dilu}. For analogy, humans can always drive a car correctly by learning driving skills and traffic rules for a few months. In other words, learning how to drive is accomplished via prior knowledge transfer without requiring a long training time. Motivated by this, knowledge-driven autonomous driving techniques \cite{li2023towards} is introduced as an attempt to address the existing difficulties of data-driven driving techniques.

In particular, recent approaches \cite{wen2023dilu,chen2023driving,contributors2023drivelm} have integrated Large Language Models(LLMs) and Vision-Language Models(VLMs) into autonomous driving, emphasizing language reasoning, decision-making, and vision-text fusion. Additionally, Multimodal Large Language Models(MLLMs) \cite{wang2023drivemlm,cui2024survey} merge natural language processing with computer vision to efficiently interpret complex traffic scenarios, including pedestrian movements, traffic signs, and vehicle behavior prediction.

MLLMs excel in processing multimodal data for autonomous driving \cite{xia2023parameterized}, achieving minimal human intervention by automating decision-making and operations \cite{chen2023end}. Their integration of visual, linguistic, and sensor data enables precise navigation and real-time decisions in complex environments \cite{kochdumper2023real}, thus achieving accurate model performances in tasks such as object detection, scene understanding, and trajectory prediction \cite{wang2023drive}.

The above findings have motivated the question: \textbf{How can we apply the capabilities of MLLMs to autonomous driving systems?} 

Training MLLMs is often complex and costly. Even with fine-tuning techniques like LORA \cite{hu2021lora}, substantial computational resources and data are required \cite{li2020survey}. In terms of developing MLLM capabilities in autonomous driving more cost-effectively, the prompt techniques has shown immense potential \cite{brown2020language,gao2020fewshot}.Research \cite{kong2023better} on motivating language models, it has been found that Zero-shot-CoT(Chain-of-Thought) can be used to generate instances of thought process at each traffic scenario through "Let's think step by step", together with an answer-directed prompt to motivate the model in generating answers \cite{wang2023plan}. This through process of Zero-shot-CoTresembles the human thought process \cite{kojima2022large}, thus better capitalizing the language-driven ability of MLLMs.

\begin{figure}
\includegraphics[width=12cm]{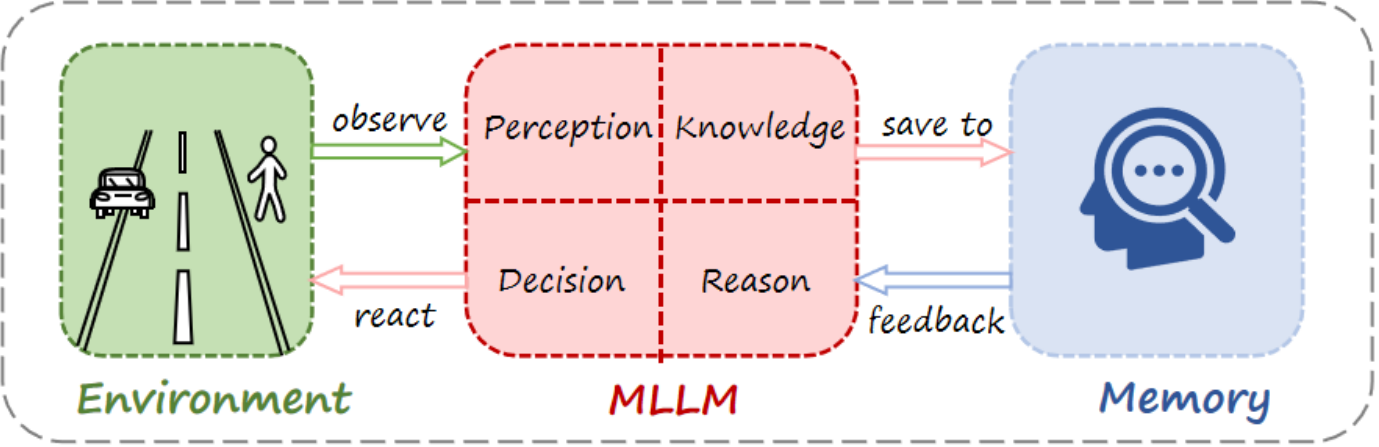}
\caption{The MLLM model operates as a driver agent within the PKRD-CoT paradigm, including a dynamic environment, an MLLM model with capabilities in perception, knowledge, reasoning, and decision-making, and a memory module that stores information in JSON format. The MLLM model continuously senses the environment, recognizes targets, reasons about situations, and interacts with the memory module to make decisions for controlling the car.} 
\label{fig1}
\end{figure}

Therefore, we have designed a Zero-shot-CoT prompt framework, named PKRD-CoT(Perception, Knowledge, Reasoning, and Decision-making Chain-of-Thought) as \textbf{Figure \ref{fig1}}, derived from four fundamental capabilities of autonomous driving, i.e., perception, knowledge, reasoning, and decision-making. Autonomous driving systems require "perception" and "knowledge" to understand their environment, while "reasoning" and "decision-making" are in respond to dynamic traffic conditions, which is crucial for their safety and efficiency, and "memory" can effectively resolve contextual continuity problems in LLM . Two experiments on the real-world Nuscene dataset and a highway simulation dataset are used to thoroughly validate the the understanding and reasoning ability of MLLMs in real driving scenarios. Our results show that MLLMs with PKRD-CoT improve end-to-end autonomous driving.

Our contributions in this study are highlighted as follows:
\begin{itemize}
\item We have designed an effective PKRD-CoT framework that incorporates a Chain-of-Thought prompting for perception, knowledge, reasoning, and decision-making, tailored to autonomous driving. It is the first work that applies prompt engineering in MLLMs to autonomous driving tasks in obtaining a more interpretable and reliable driving decision. 

\item The results of the ablation experiments show that our PKRD-CoT framework outperforms traditional zero-shot and role-playing prompts, with a 22\% and 6\% improvement in decision making accuracy respectively.

\item We propose an evaluation system for assessing the suitability of MLLMs for autonomous driving tasks, based on PKRD-CoT framework. This evaluation results show that GPT-4 achieves the overall best MLLM results, followed by Claude and LLava 1.6. In particular, CogVLM excels in target localization and perception tasks, while surprisingly, Minigpt-4 performs poorly in mathematical reasoning tasks. These findings provide crucial insights for the advancement of autonomous driving technologies.
\end{itemize}

\section{Related Work}
\subsection{Multi-Modal Large Language Models}
The rapid evolution of Language Large Models (LLMs) has continuously been improving their abilities to understand the world \cite{zhang2024mm}. Multimodal Language Large Models (MLLMs), as an extension of LLMs, represent a more advanced level, enabling language understanding that is no longer confined to text, but capable of comprehensively understanding and generating multiple types of data, including non-textual data such as images, audio and video \cite{cui2024survey,yin2023survey}.

Recent advances in computational power and data availability have spurred the development of multimodal macro models, enhancing complex content understanding and task precision. Notable models like GPT4.0 \cite{koubaa2023gpt}, which is known for its broad application success and data handling; MiniGPT-4\cite{zhu2023minigpt}, a streamlined version of GPT4.0 with efficient linguistic processing and lower resource needs; InstructBLIP \cite{dai2024instructblip} and LLaVA \cite{touvron2023llama} for rapid image recognition and linguistic output; Qwen-VL \cite{bai2023qwen} for image localization and mathematical tasks; CogVLM \cite{wang2023cogvlm} for image comprehension and dialogue; and Anthropic's Claude AI \cite{anthropic2024claude}, excelling in reasoning, math, and code generation.

Currently, more studies are using MLLMs for autonomous driving \cite{wen2023dilu,shao2023lmdrive,yang2023llm4drive,xu2023drivegpt4,wang2023bevgpt}. A notable example of the autonomous driving systems is DriveMLM \cite{wang2023drivemlm}, which combines the MLLM's decision-making outputs with the behavioral planning phase of autonomous driving, enabling the driving system to understand complex driving scenarios and make smart decisions. 

A MLLM-integrated autonomous driving system will be more accurate, interpretable and safe. On this account, applying these various MLLM capabilities, such as sensing, recognizing, reasoning, and decision-making, is an essential pressing issue of autonomous driving.

\subsection{Language Prompt}
Prompt engineering has emerged as a critical technique in adapting MLLMs to specific tasks and environments\cite{liu2021pretrain}. Prompt engineering techniques take many forms, including zero-sample prompts, few-sample prompts, role-playing prompts, and chain-of-thought prompts\cite{li2023efficient}\cite{zhang2023knowledge}.

Among the various types of prompts, Zero-Shot Chain of Thought (CoT) prompts have shown remarkable effectiveness in guiding MLLMs. Kojima et al. \cite{kojima2022zero} highlighted the Zero-Shot CoT prompting method, showing its potential to improve reasoning and decision-making without requiring additional task-specific training data \cite{wei2022chain}.

Zero-shot Chain of Thought (Zero-shot-CoT) enables complex question answering across domains without task-specific model training by simulating human problem-solving thought processes in language models \cite{kong2023better}. It improves the model's adaptability, flexibility, and explanatory power \cite{qin2023cross}, especially when dealing with abstract and multi-step reasoning tasks.

According to the studies, the models applying Zero-shot-CoT demonstrated significant performance improvement on multiple standard datasets \cite{voge2024leveraging}. Using Zero-shot-CoT, models significantly improved multi-step reasoning accuracy \cite{wang2023plan} and demonstrated enhanced understanding and answer generation in complex reading comprehension and question-answering tasks \cite{kojima2022large}.

The Zero-shot-CoT is crucial for fully realizing the potential of MLLMs for end-to-end autonomous driving. Through these unique advantages, Zero-shot-CoT guides the model to find the optimal solution in its thought process \cite{kong2023better} to facilitate a smooth end-to-end process from perception to decision-making for autonomous driving. 

\section{PKRD-CoT}
\subsection{PKRD-CoT framework}
PKRD-CoT is a specialized prompt framework designed for autonomous driving tasks. It derives its capabilities from four fundamental aspects: perception, knowledge, reasoning, and decision-making, ensuring safety and efficiency in dynamic traffic conditions. The process is carried out step by step through Observation, Identification, Memory, and Decision, which we detail in the following.

The first step is to observe the environment and form a current understanding of the surroundings. It ensures that the system accurately perceives relevant details and dynamics of the driving environment.

The second step is to identify and predict specific targets within the environment. It ensures that the system can focus on and track important elements, such as other vehicles and pedestrians, crucial for safe navigation.

The third step memory stores the environmental understanding in a structured JSON format, following a designed template. This module addresses the limitations of language models in maintaining context over extended interactions, enabling the system to retain and utilize past information effectively.

The last step is to make informed decisions based on the current understanding of the present information and the stored information from the memory. It ensures that the system can respond appropriately to dynamic traffic conditions.

\subsection{PKRD-CoT Performance in GPT4.0}
This section explores and validates whether GPT-4.0 can perform autonomous driving tasks within the PKRD-CoT framework.

\underline{Environment Construction}: 
In the environment construction, we select a subset of data from a public autonomous driving datasets, Nuscenes \cite{caesar2020nuscenes}, which comprises complex and diverse driving scenarios to conduct tests and experiments. In particular, we merge the images from six cameras in Nuscenes, right-front, front, and left-front images are merged into the front image; right-back, back, and left-back are merged into the back image, to form a panoramic environment around the car. This panoramic view not only enhances GPT4.0's overall awareness of the surrounding context, but also effectively reduces the long tail problem (weak understanding and generation for rare or uncommon scenarios) existing in GPT4.0.

\underline{Input and Output}: 
We input the panoramic view of the car's environment (front and back view formed by 6 images) alongside the PKRD-CoT prompt. In doing this, we expect GPT-4.0 to utilize its Scene Perception, Object Detection, Localization, and Semantic Understanding modules to execute the task and ultimately provide the analysis and the answer to the task.

\begin{figure}[H]
  \centering
    \includegraphics[width=9cm]{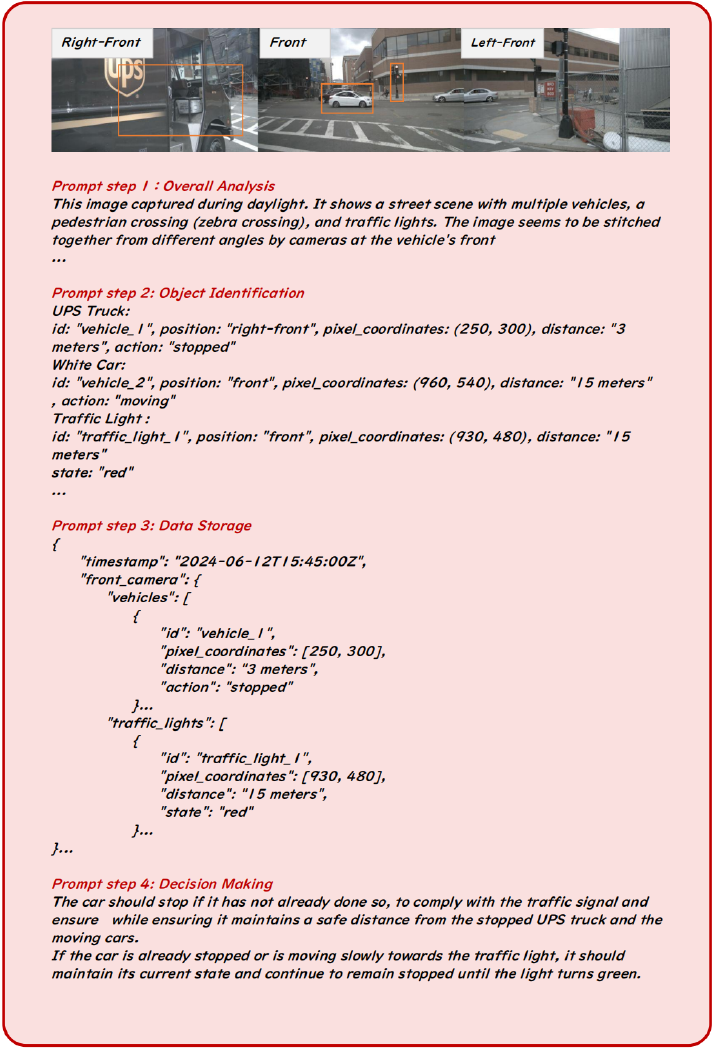}
  \caption{Example Outputs of GPT4.0 with PKRD-CoT in Autonomous Driving}
  \label{fig2}
\end{figure}

In our experiments, we tested our proposed PKRD-CoT on three scenarios, which includes highway, daytime traffic and nighttime traffic, focusing on cars, pedestrians, traffic signs (traffic lights and zebra crosswalks), and the environment around the car (daytime, nighttime, or rainy or sunny days). For object detection in image space of the cars, pedestrians, and traffic signs (traffic lights and crosswalks), we designed five attributes for organizing them: "ID", "Position", "Pixel-Coordinates (the pixel coordinates of the object in the image)", "State" (the observed status of the object).

After obtaining this representation of the environmental scene, we designed five proceeded to input selection prompts of driving behaviors: speed up, speed down, stop, keep remain or change lane, as the input selection prompts for GPT4.0 to make initial driving decisions. The outputs of GPT4.0 according to our prompt is shown in the following \textbf{Figure \ref{fig2}}.

\underline{Experimental Analysis}: 
From our experiments, we found out that GPT4.0 has superior performances in scene perception, object detection, and semantic comprehension, which we detail in the following.

\textbf{Full Observation}: GPT4.0 accurately understands and describes the content of the scene based on the contextual information of the given scenario in the image, identifying the main elements and actions and the relationship between them. For example, given the scenario of a cross-traffic light intersection, GPT4.0 conducts a preliminary analysis and quickly locates the current scenario as belonging to be "during daytime", "a street scene with multiple vehicles, a pedestrian crossing (zebra crossing), and traffic lights", and the camera position of the image is also predicted that "The image seems to be composed of six different sections, probably from cameras located at the front and back of a vehicle, showing different angles".

\textbf{Knowledge Driven}: GPT4.0 is able to capture overall information after image and semantic alignment. In particular, its perception of the target goes beyond simple object identification; it can understand and make decisions using its extensive knowledge. For instance, when it detects an object in front of a "Traffic Light: Red," it can "think" like a human driver and respond appropriately via "Slow Down" or "Stop". 

\subsection{PKRD-CoT Ablation Study}
To validate the PKRD-CoT prompt framework, we designed and implemented a series of ablation experiments. These experiments are conducted on images of real driving scenes, and GPT4.0's final decision is used subjected to evaluation. The accuracy of the this ablation study is evaluated as the number of correct decisions divided by the total number of samples. These experiments were designed to assess the effects of different prompt engineering methods: zero-shot, role-playing, and PKRD-CoT on the performance of the MLLM model, i.e., GPT4.0 in our case.

Prompt Zero-Shot: we defined a baseline experimental condition in which the model was allowed to make driving decisions without any cues, relying solely on its built-in capabilities for decision-making and task execution without additional auxiliary information or structural prompts. 

Prompt Role-Playing: we introduced role-playing prompts that simulate specific scenarios and roles (intelligent drivers of self-driving cars) to stimulate the model's ability to respond more naturally and intelligently in the simulated environment. 

Prompt PKRD-CoT: By explicitly structuring the chain of thought, and based on the four important capabilities required for autonomous driving, PRKD-CoT helps the model to decompose complex tasks step by step, ensuring that each step of the reasoning process is interpretable, thus ultimately improves the accuracy and reliability of task execution.

The experimental results in \textbf{Table \ref{tab:ablation_experiment}} show that, compared to the baseline condition (zero-shot), role-play prompts slightly improve the model's task completion rate and decision accuracy. However, the PKRD-CoT framework significantly outperforms both the baseline and role-play methods, demonstrating its superior effectiveness and utility for autonomous driving tasks. This verifies the strong potential of the PKRD-CoT framework in enhancing model performance in complex, dynamic environments.

\begin{table}[ht]
\centering
\caption{Ablation Experiment Results, showing the accuracy of different prompt engineering methods in decision-making using 100 real scene samples. The zero-shot method achieved 72\% accuracy, the role-playing method achieved 88\% accuracy, and the PKRD-CoT method achieved the highest accuracy at 94\%, demonstrating the effectiveness of the PKRD-CoT framework.}
\label{tab:ablation_experiment}
\begin{tabular}{lc}
\toprule
\textbf{Method} & \textbf{Accuracy} \\
\midrule
GPT-4.0 + Zero-Shot & 72\% \\
GPT-4.0 + Role-Playing & 88\% \\
GPT-4.0 + PKRD-CoT & 94\% \\
\bottomrule
\end{tabular}
\end{table}

\begin{figure}[h]
  \centering
    \includegraphics[width=10cm]{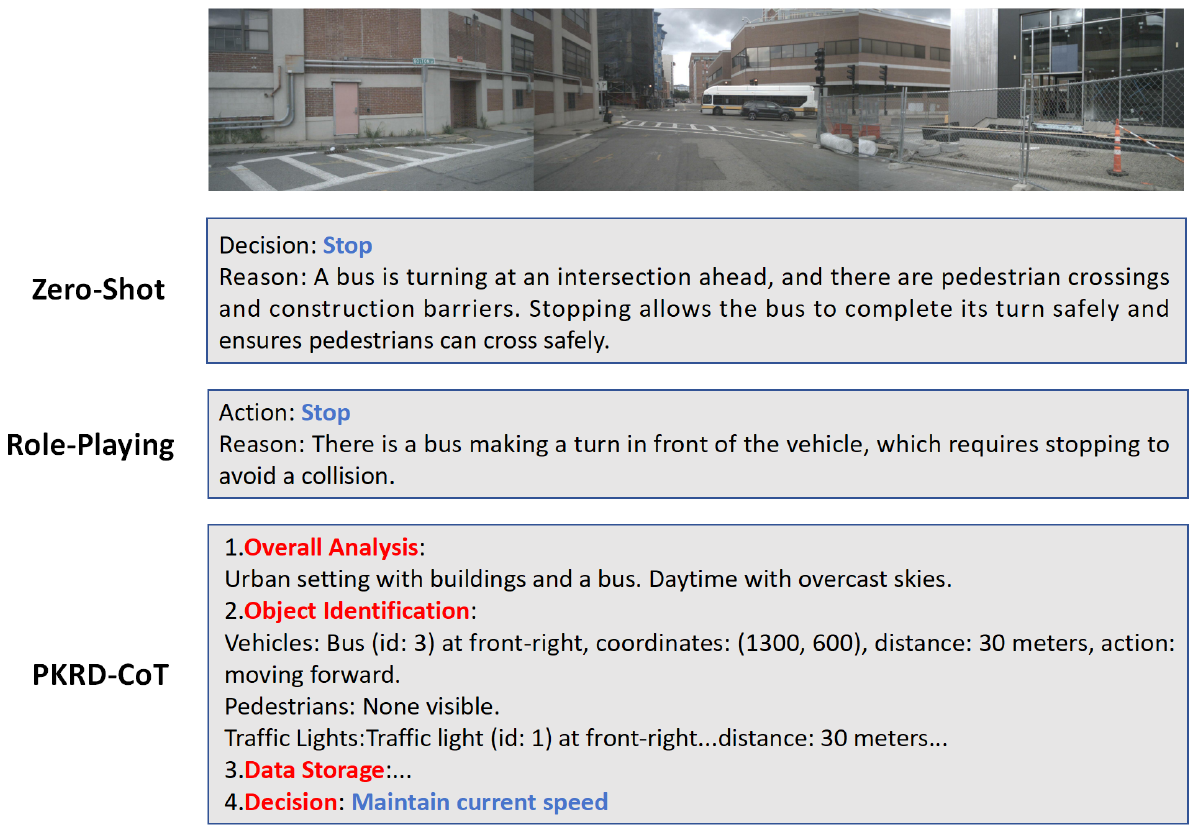}
  \caption{Example of Ablation Experiment Outputs}
  \label{fig3}
\end{figure}

For example, \textbf{Figure \ref{fig3}} shows the different decisions and reasons provided by these three different prompt methods. The results highlights the superiority of PKRD-CoT, where it decides to maintain speed in the given scenario. Maintaining speed is in fact a more appropriate decision, as opposed to stopping suggested by zero-shot and role-playing. This demonstrates a more precise understanding in physical distances and its capability of making optimal decision.

\section{MLLMs Capabilities Evaluation Experiments}
In the previous section, we has demonstrated that by incorporating the PKRD-CoT prompt, GPT-4.0 is able to achieve high-fidelity decision making for self-driving. Building on these experimental insights, we further explored the integration of other MLLMs with our PKRD-CoT prompt framework and assessed their performance in autonomous driving.

Given that the essence of our PKRD-CoT lies in four essential capabilities for autonomous driving, we conducted experiments to evaluate various models based on these key abilities: perception, knowledge, reasoning, and decision-making. Here, we selected six representative state-of-the-art MLLMs:
GPT4.0 \cite{koubaa2023gpt}, claude \cite{anthropic2024claude}, LLava1.6 \cite{touvron2023llama}, Qwen-VL-Plus \cite{bai2023qwen}, CogVLM chat \cite{wang2023cogvlm}, and Minigpt4 \cite{zhu2023minigpt}, for comparisons.

\subsection{Perception Ability}

\begin{table}[htbp]
\centering
\caption{Results of MLLMs perceptual accuracy, showing the perceptual capabilities of MLLMs using real image samples on five evaluation metrics: cars, crowds, traffic lights, crosswalks, and the current scene. The model's accuracy is 100\% if it correctly identifies and describes the target, otherwise, it's 0\%. The average accuracy for each category and the overall average accuracy are used to assess the model's performance.}
\label{tab1}
\begin{tabular}{lcccccc}
\toprule
\textbf{Model} & \textbf{Car} & \textbf{People} & \textbf{\shortstack{Traffic \\ Light}} & \textbf{\shortstack{Pedestrian \\ Crossing}} & \textbf{\shortstack{Current \\ Scene}} & \textbf{Average} \\
\midrule
GPT-4.0\cite{koubaa2023gpt} & 100\% & 90\% & 100\% & 90\% & 100\% & 96\% \\
Claude\cite{anthropic2024claude} & 100\% & 90\% & 90\% & 90\% & 100\% & 94\% \\
LLava1.6\cite{touvron2023llama} & 100\% & 90\% & 90\% & 80\% & 100\% & 92\% \\
Qwen-VL-Plus\cite{bai2023qwen} & 100\% & 100\% & 40\% & 90\% & 100\% & 86\% \\
CogVLM chat\cite{wang2023cogvlm} & 100\% & 77.78\% & 55.56\% & 66.67\% & 66.67\% & 73.34\% \\
Minigpt4\cite{zhu2023minigpt} & 100\% & 55.56\% & 44.44\% & 89\% & 66.67\% & 72.22\% \\
\bottomrule
\end{tabular}
\end{table}

\begin{figure}[h]
  \centering
    \includegraphics[width=10cm]{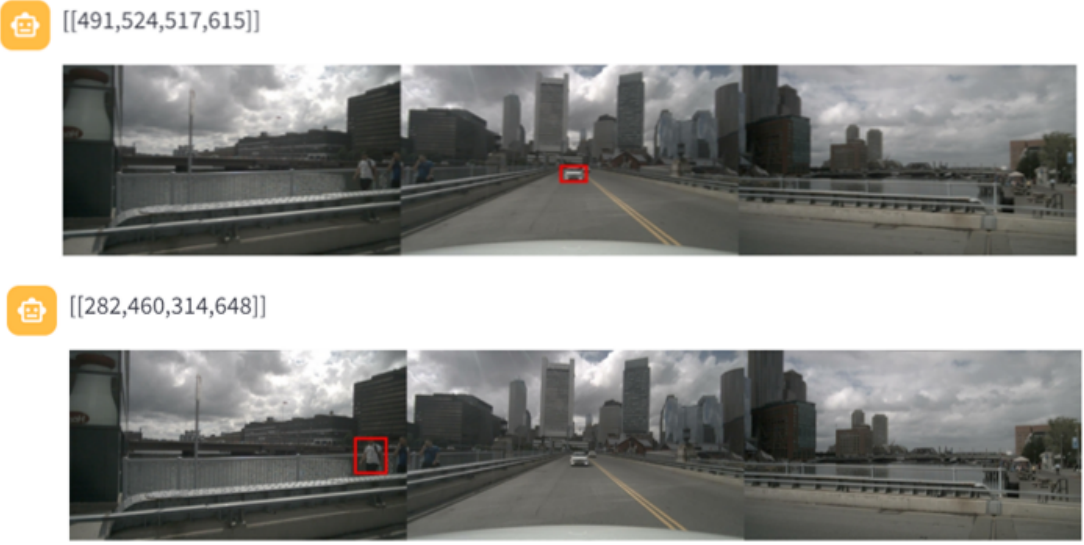}
  \caption{CogVLM chat: Positioning of Car and Pedestrian}
  \label{fig4}
\end{figure}

We obtained real images from public datasets and tested them on the following five target species: Car, People, Traffic Lights, Pedestrian Crossings, and Current Scene. The evaluation criteria are based on the model's ability to accurately identify the target species and generate the corresponding descriptions: if the model correctly recognizes and describes the target, the perceptual accuracy for that target species is taken to be 100\%; ; conversely, if the recognition is incorrect, the perceptual accuracy is 0\%. The average perceptual accuracy for each target species is calculated from these samples, providing the final perceptual accuracy for that species. Furthermore, we calculated the overall model's perceptual accuracy by averaging the perceptual accuracy across the five target categories.

From the quantitative results in \textbf{Table \ref{tab1}}, we observed that all models achieve 100\% perceptual accuracy in the "Car" category, indicating satisfactory driving scene understanding. For "People," Qwen-VL-Plus performs best with 100\% accuracy. For "Traffic Light", most models, especially Qwen-VL-Plus and CogVLM chat, perform poorly, which is likely due to the diversity of traffic lights and varying lighting conditions. Recognizing "Pedestrian Crossing" requires understanding traffic rules and human behaviors, and all models except for CogVLM chat and LLava1.6 perform well. For "Current Scene", all models except Minigpt4 and CogVLM chat achieve 100\% accuracy, demonstrating overall a good scene understanding.

Nevertheless, GPT4.0 performs the best in these experiments, showing its superior perceptual performance across all the target species. The performances of Claude and LLava1.6 come close to GPT4.0 with average perceptual accuracies of 94\% and 92\%. Qwen-VL-Plus has average perceptual accuracies that are slightly lower than the previous two models. It is worth noting that Qwen-VL-Plus performs well in sensing other targets except for "Traffic Light". During the experiments, we also noticed that the CogVLM chat has unique advantages in specific scenarios. Despite not attaining the highest accuracy, it has the best performance in localizing the target, as shown in \textbf{Figure \ref{fig4}}.

\subsection{Knowledge Ability}

\begin{figure*}[h]
\centering
\includegraphics[width=12cm]{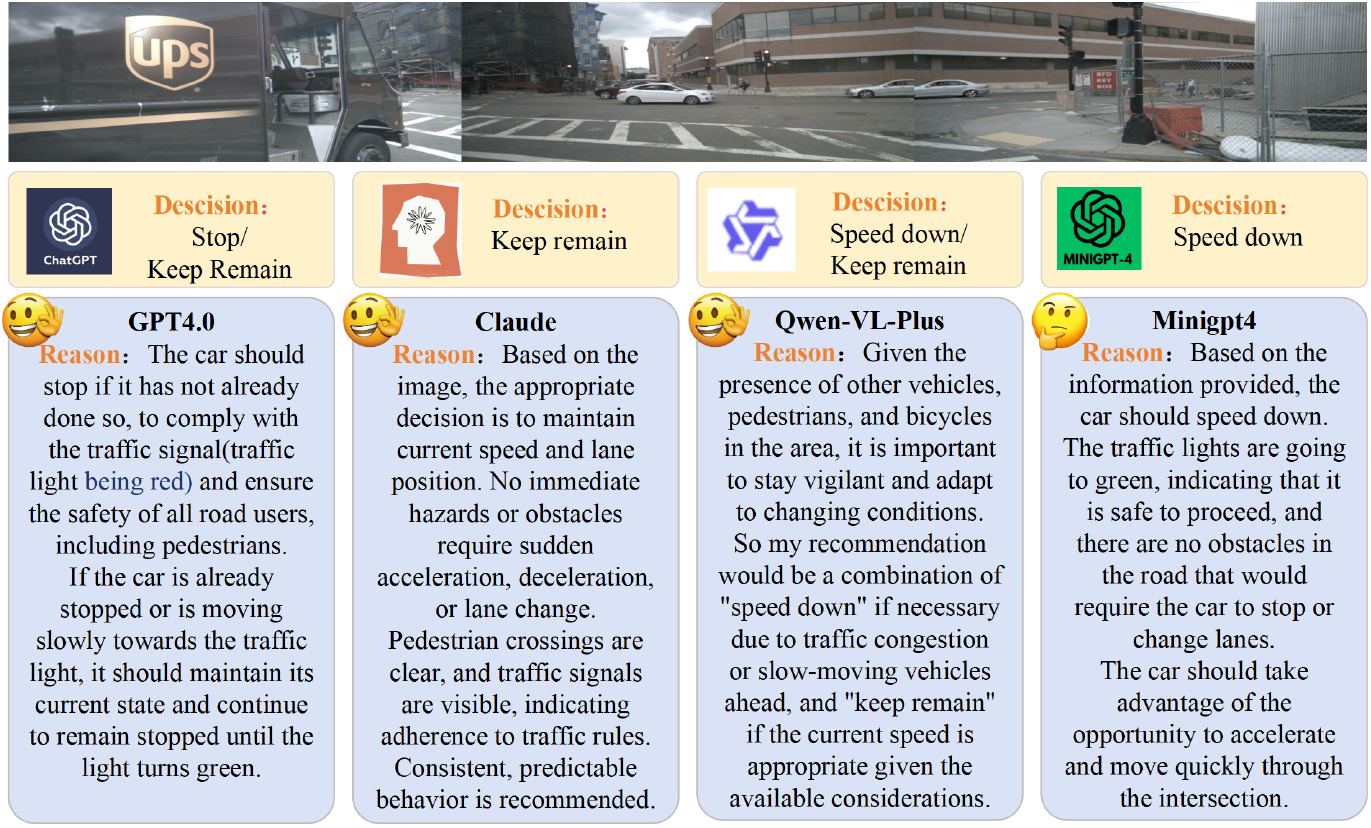}
\caption{Visual Results of Knowledge Ability Comparative Experiments}
\label{fig5}
\end{figure*}

We further tested these MLLMs on the panoramic view of the car, each MLLM will act as an intelligent driver and attempt to accomplish identified tasks. The models scrutinize and analyze the driving scenarios based on the given driving scenarios information and their own knowledge. \textbf{Figure \ref{fig5}} shows an example of the models' decision-making a given scenario.

In conclusion, CogVLM chat, GPT4.0, Claude, and LLava 1.6 show emphasis on safety by tending towards stopping or remaining the current speed at red lights and crosswalks. This suggests a cognitive ability that includes understanding and adhering to traffic rules. On the contrary, Minigpt4 is inclined towards utilizing information of upcoming signal changes for making decisions, thus showing an ability to predict the environment but putting less emphasis on safety, which is essential in real-world traffic environments. GPT4.0, on the other hand, takes into account the dynamics of the incoming red light and the vehicles present at the moment, to make a more accurate and informed driving decision. Both of GPT4.0 and Qwen-VL-Plus show the model's ability to adapt to a dynamic environment.
Qwen-VL-Plus considers not only traffic rules but also pedestrians, showing its complexity and comprehensiveness in cognitive processing.

\subsection{Mathematical Reasoning Ability}
In addition, the ability to perform mathematical calculations is crucial for scene understanding in autonomous driving, such as calculating relative speed and maintaining a safe distance between vehicles. Therefore, we assessed mathematical computation as a measure of reasoning ability.

To test the mathematical ability of the model, we conducted a computational experiment during driving, using the safe vehicle distance as the reference object. MLLMs calculate the vehicle distance using the coordinates of the vehicle and the Pythagorean theorem formula. The calculated distance is then compared with the ground-truth distance, allowing for calculating the error percentage, which will be used to verify the model's computing ability.

This experiment proved that the model can automatically evaluate the Pythagorean theorem for distance calculation with varying parameters, even without additional prompts. A model was counted as correct if it could correctly compute the distance with an absolute error of not more than 0.5m. Then, the percentage of correct model calculations is computed to evaluate the model's computational ability. The results are shown in the \textbf{Table \ref{tab2}}.

\begin{table}[ht]
\centering
\caption{Mathematical Calculation Ability of MLLMs, showing the computational power of MLLMs with simulated driving scenarios where the model computes distances between vehicles. A calculation is correct if the absolute error is not more than 0.5m. The correct rate and the percentage of accurate calculations are used to assess the model's computational ability.}
\label{tab2}
\begin{tabular}{lcc}
\toprule
\textbf{Model} & \textbf{\shortstack{Pythagorean \\ Theorem}} & \textbf{Correct} \\
\midrule
GPT4.0\cite{koubaa2023gpt} & $\sqrt{}$ & 100\% \\
Claude\cite{anthropic2024claude} & $\sqrt{}$ & 90\% \\
LLava1.6\cite{touvron2023llama} & $\sqrt{}$ & 80\% \\
Qwen-VL-Plus\cite{bai2023qwen} & $\sqrt{}$ & 30\% \\
CogVLM chat\cite{wang2023cogvlm} & $\sqrt{}$ & 20\% \\
Minigpt4\cite{zhu2023minigpt} & - & - \\
\bottomrule
\end{tabular}
\end{table}

The results show that GPT-4.0 demonstrates excellent mathematical reasoning capabilities, achieving 100\% accuracy. This result shows its efficiency in logical reasoning and algorithm implementation. Claude and LLava1.6 also show relatively high computational accuracy of 90\% and 80\%, respectively, showing their reliability in performing mathematical computing tasks. Qwen-VL-Plus and CogVLM chat have significantly lower correctness rates, at only 30\% and 20\%. Minigpt4 did not have the computational capability of the Pythagorean theorem in this evaluation. It is able to call the formula, but fail to evaluate the correct values.

\subsection{Decision-Making Ability}
In autonomous driving, perception, knowledge, and reasoning are used to inform the final decision. Decision-making capability integrates the vehicle's perception, knowledge, environmental changes, and predicted future trends to make optimal choices. The driving agent needs to quickly process large amounts of data, while demonstrating high adaptability and flexibility to respond to the complex and changing driving environment.

\begin{figure*}[h]
\includegraphics[width=12cm]{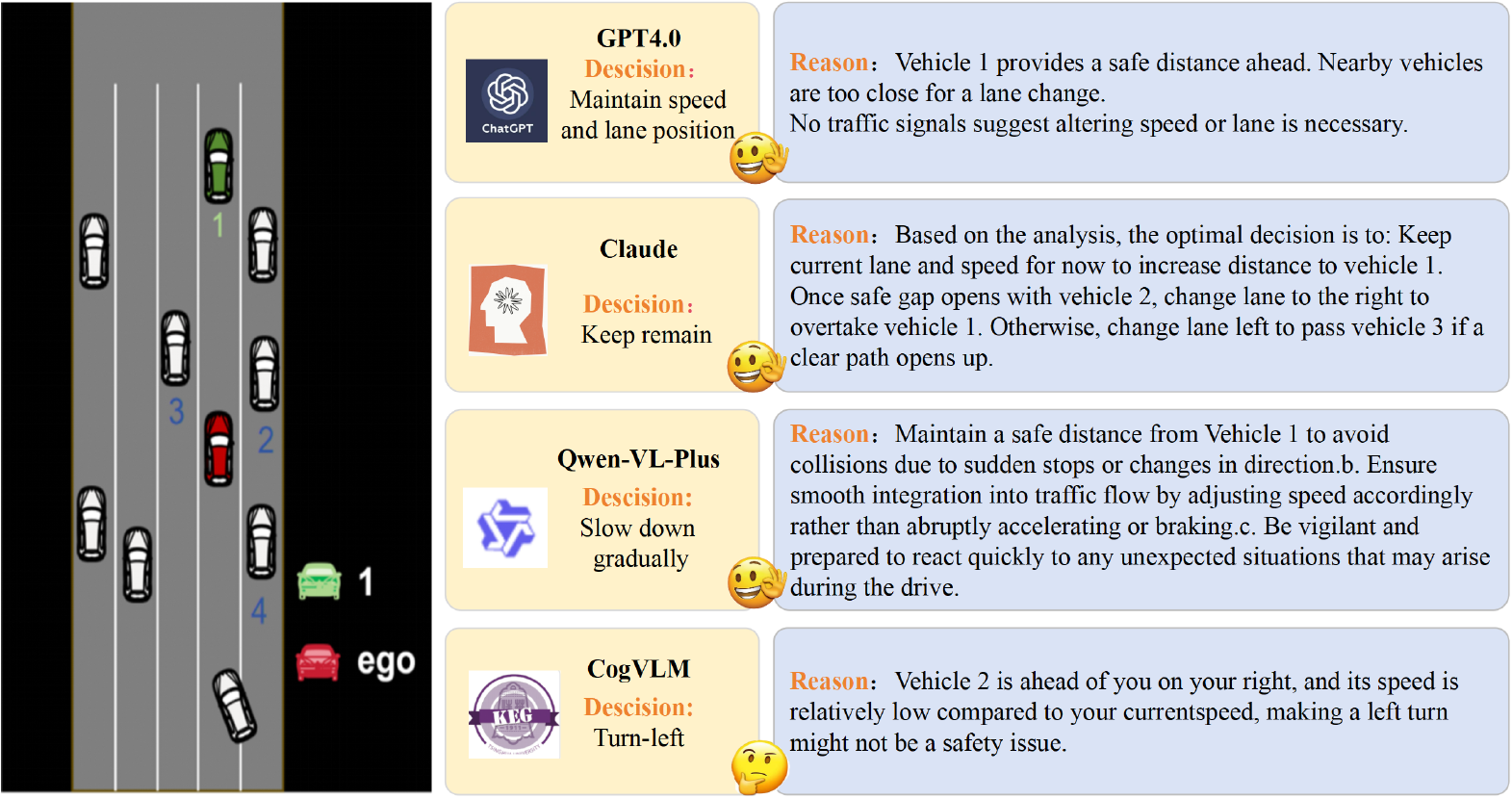}
\caption{Visual Results of Decision-Making Ability Comparative Experiments}
\label{fig6}
\end{figure*}

Here, we choose highway as the experimental scenario and used the BEV images to assess the decision-making ability of the MLLMs on the highway environment. To achieve this, we first convert driving scenario data to text for informed input, and then input our PKRD-CoT prompts to guide the output decisions of these models. Outputs of the MLLMs are shown in the \textbf{Figure \ref{fig6}}.

Based on the results, GPT-4.0, Claude, and LLava1.6 adopt a cautious approach to decision-making, preferring to maintain the current state until a significant need for change arises. This method is effective in stable traffic conditions as it minimizes risks, albeit sacrificing adaptability and foresight. In contrast, Qwen-VL-Plus prioritizes smooth speed adjustments for safety and passenger comfort, thus demonstrating greater flexibility and potential effectiveness in unexpected situations. Meanwhile, CogVLM chat and Minigpt4 favor proactive lane changes, which highlights their ability for swift decision-making. However, this strategy may introduces additional risks if the models inaccurately judge speed and distance.

\section{Conclusion}
Our research presents an innovative PKRD-CoT framework that combines chain-of-mind cues with the essential capabilities (perception, knowledge, reasoning, and decision-making) essential for autonomous driving. This marks the first application of prompt engineering to multimodal large language models (MLLMs) for an autonomous driving task. The seamless integration of these capabilities within the PKRD-CoT framework allows MLLMs to perform knowledge-driven driving without the need for pre-training.

In addition, we present a set of comprehensive evaluation tests to assess the suitability of various MLLMs for autonomous driving tasks, focusing on the four key driving capabilities. Our results show that GPT-4 outperforms the other models overall, with Claude and LLava 1.6 following in second place. In particular, CogVLM excels in target localization and perception tasks, while MiniGPT-4 shows weaker performance in mathematical reasoning. These tests offer valuable insights into the strengths and limitations of the different MLLMs, crucial for future advancements in LLM-based autonomous driving systems.

In conclusion, our study introduces a novel and effective framework that leverages PKRD-CoT in conjunction with MLLMs for autonomous driving. This unified approach enhances real-time decision-making and fosters further innovation in the field. Potential future work in this area includes refining knowledge integration and exploring advanced reasoning capabilities to further improve the safety and efficiency of autonomous driving systems.

\bibliographystyle{splncs04}
\bibliography{references}

\begin{thebibliography}{10}
\providecommand{\url}[1]{\texttt{#1}}
\providecommand{\urlprefix}{URL }
\providecommand{\doi}[1]{https://doi.org/#1}

\bibitem{anthropic2024claude}
Anthropic: The claude 3 model family: Opus, sonnet, haiku. Anthropic  (2024), online

\bibitem{bai2023qwen}
Bai, J., et~al.: Qwen-vl: A frontier large vision-language model with versatile abilities. arXiv preprint arXiv:2308.12966  (2023)

\bibitem{bogdoll2021description}
Bogdoll, D., et~al.: Description of corner cases in automated driving: Goals and challenges. In: Proceedings of the IEEE/CVF International Conference on Computer Vision. pp. 1023--1028 (2021)

\bibitem{bolte2019towards}
Bolte, J.A., et~al.: Towards corner case detection for autonomous driving. In: 2019 IEEE Intelligent vehicles symposium (IV). pp. 438--445 (2019)

\bibitem{brown2020language}
Brown, T., et~al.: Language models are few-shot learners. In: Advances in neural information processing systems. vol.~33, pp. 1877--1901 (2020)

\bibitem{caesar2020nuscenes}
Caesar, H., et~al.: nuscenes: A multimodal dataset for autonomous driving. In: Proceedings of the IEEE/CVF Conference on Computer Vision and Pattern Recognition. pp. 11621--11631 (2020)

\bibitem{chen2023end}
Chen, L., et~al.: End-to-end autonomous driving: Challenges and frontiers. arXiv preprint arXiv:2306.16927  (2023)

\bibitem{chen2023driving}
Chen, L., et~al.: Driving with llms: Fusing object-level vector modality for explainable autonomous driving. arXiv preprint arXiv:2310.01957  (2023)

\bibitem{contributors2023drivelm}
Contributors: Drivelm: Drive on language  (2023)

\bibitem{cui2024survey}
Cui, C., et~al.: A survey on multimodal large language models for autonomous driving. In: Proceedings of the IEEE/CVF Winter Conference on Applications of Computer Vision. pp. 958--979 (2024)

\bibitem{dai2024instructblip}
Dai, W., et~al.: Instructblip: Towards general-purpose vision-language models with instruction tuning. Advances in Neural Information Processing Systems  \textbf{36} (2024)

\bibitem{gao2020fewshot}
Gao, T., et~al.: Making pre-trained language models better few-shot learners. arXiv preprint arXiv:2012.15723  (2020)

\bibitem{hu2021lora}
Hu, E., et~al.: Lora: Low-rank adaptation of large language models. arXiv preprint arXiv:2106.09685  (2021)

\bibitem{kochdumper2023real}
Kochdumper, N., Bak, S.: Real-time capable decision making for autonomous driving using reachable sets. arXiv preprint arXiv:2309.12289  (2023)

\bibitem{kojima2022zero}
Kojima, T., et~al.: Large language models are zero-shot reasoners. arXiv preprint arXiv:2205.11916  (2022)

\bibitem{kojima2022large}
Kojima, T., et~al.: Large language models are zero-shot reasoners. In: Advances in neural information processing systems. vol.~35, pp. 22199--22213 (2022)

\bibitem{kong2023better}
Kong, A., et~al.: Better zero-shot reasoning with role-play prompting. arXiv preprint arXiv:2308.07702  (2023)

\bibitem{koubaa2023gpt}
Koubaa, A.: Gpt-4 vs. gpt-3.5: A concise showdown  (2023)

\bibitem{li2023efficient}
Li, B., Li, F., Gao, S., Fan, Q., Lu, Y., Hu, R., Zhao, Z.: Efficient prompt tuning for vision and language models. In: International Conference on Neural Information Processing. pp. 77--89. Springer (2023)

\bibitem{li2020survey}
Li, P., et~al.: A survey on deep learning-based approaches for autonomous driving. IEEE Transactions on Intelligent Transportation Systems  (2020)

\bibitem{li2023towards}
Li, X., et~al.: Towards knowledge-driven autonomous driving. arXiv preprint arXiv:2312.04316  (2023)

\bibitem{liu2021pretrain}
Liu, P., et~al.: Pre-train, prompt, and predict: A systematic survey of prompting methods in natural language processing. arXiv preprint arXiv:2107.13586  (2021)

\bibitem{qin2023cross}
Qin, L., et~al.: Cross-lingual prompting: Improving zero-shot chain-of-thought reasoning across languages. arXiv preprint arXiv:2310.14799  (2023)

\bibitem{shao2023lmdrive}
Shao, H., et~al.: Lmdrive: Closed-loop end-to-end driving with large language models. arXiv preprint arXiv:2312.07488  (2023)

\bibitem{touvron2023llama}
Touvron, H., et~al.: Llama 2: Open foundation and fine-tuned chat models. arXiv preprint arXiv:2307.09288  (2023)

\bibitem{voge2024leveraging}
{V\"o}ge, L., et~al.: Leveraging zero-shot prompting for efficient language model distillation. arXiv preprint arXiv:2403.15886  (2024)

\bibitem{wang2023plan}
Wang, L., et~al.: Plan-and-solve prompting: Improving zero-shot chain-of-thought reasoning by large language models. arXiv preprint arXiv:2305.04091  (2023)

\bibitem{wang2023bevgpt}
Wang, P., et~al.: Bevgpt: Generative pre-trained large model for autonomous driving prediction, decision-making, and planning. arXiv preprint arXiv:2310.10357  (2023)

\bibitem{wang2023drive}
Wang, T.H., et~al.: Drive anywhere: Generalizable end-to-end autonomous driving with multi-modal foundation models. arXiv preprint arXiv:2310.17642  (2023)

\bibitem{wang2023cogvlm}
Wang, W., et~al.: Cogvlm: Visual expert for pretrained language models. arXiv preprint arXiv:2311.03079  (2023)

\bibitem{wang2023drivemlm}
Wang, W., et~al.: Drivemlm: Aligning multi-modal large language models with behavioral planning states for autonomous driving. arXiv preprint arXiv:2312.09245  (2023)

\bibitem{wei2022chain}
Wei, J., Wang, X., Schuurmans, D., Bosma, M., Chi, E.H., Le, Q.V., et~al.: Chain of thought prompting elicits reasoning in large language models. arXiv preprint arXiv:2201.11903  (2022)

\bibitem{wen2023dilu}
Wen, L., et~al.: Dilu: A knowledge-driven approach to autonomous driving with large language models. arXiv preprint arXiv:2309.16292  (2023)

\bibitem{xia2023parameterized}
Xia, Y., et~al.: Parameterized decision-making with multi-modal perception for autonomous driving. arXiv preprint arXiv:2312.11935  (2023)

\bibitem{xu2023drivegpt4}
Xu, Z., et~al.: Drivegpt4: Interpretable end-to-end autonomous driving via large language model. arXiv preprint arXiv:2310.01412  (2023)

\bibitem{yang2023llm4drive}
Yang, Z., et~al.: Llm4drive: A survey of large language models for autonomous driving. arXiv e-prints arXiv:2311  (2023)

\bibitem{yin2023survey}
Yin, S., et~al.: A survey on multimodal large language models. arXiv preprint arXiv:2306.13549  (2023)

\bibitem{zhang2024mm}
Zhang, D., et~al.: Mm-llms: Recent advances in multimodal large language models. arXiv preprint arXiv:2401.13601  (2024)

\bibitem{zhang2023knowledge}
Zhang, L., Li, R.: Knowledge prompting with contrastive learning for unsupervised commonsenseqa. In: International Conference on Neural Information Processing. pp. 27--38. Springer (2023)

\bibitem{zhu2023minigpt}
Zhu, D., et~al.: Minigpt-4: Enhancing vision-language understanding with advanced large language models. arXiv preprint arXiv:2304.10592  (2023)

\end{thebibliography}
\end{document}